\documentclass[letterpaper]{article} % DO NOT CHANGE THIS
\usepackage{aaai2026}
\usepackage{times}  % DO NOT CHANGE THIS
\usepackage{helvet}  % DO NOT CHANGE THIS
\usepackage{courier}  % DO NOT CHANGE THIS
\usepackage[hyphens]{url}  % DO NOT CHANGE THIS
\usepackage{graphicx} % DO NOT CHANGE THIS
\usepackage{natbib}  % DO NOT CHANGE THIS AND DO NOT ADD ANY OPTIONS TO IT
\usepackage{caption} % DO NOT CHANGE THIS AND DO NOT ADD ANY OPTIONS TO IT
\usepackage{algorithm}
\usepackage{algorithmic}

\usepackage{newfloat}
\usepackage{listings}
\usepackage{multirow}
\usepackage{booktabs}
\usepackage{longtable}
\usepackage{xcolor}
\usepackage{lipsum}
\usepackage{tcolorbox}
\usepackage[table]{xcolor}
\usepackage{diagbox}
\usepackage{subfig}
\usepackage{graphicx}
\DeclareCaptionStyle{ruled}{labelfont=normalfont,labelsep=colon,strut=off} % DO NOT CHANGE THIS
\floatstyle{ruled}
\newfloat{listing}{tb}{lst}{}
\floatname{listing}{Listing}
\title{ESG-Bench: Benchmarking Long-Context ESG Reports \\
for Hallucination Mitigation}
\author{
    Siqi Sun\equalcontrib, Ben Peng Wu\equalcontrib, Mali Jin, Peizhen Bai, Hanpei Zhang, Xingyi Song \\
}
\affiliations{
    School of Computer Science, University of Sheffield, Sheffield, UK\\
    \{siqi.sun, bpwu1, x.song\}@sheffield.ac.uk \\
}

\usepackage{bibentry}
\begin{document}

\maketitle

\begin{abstract}

As corporate responsibility increasingly incorporates environmental, social, and governance (ESG) criteria, ESG reporting is becoming a legal requirement in many regions and a key channel for documenting sustainability practices and assessing firms’ long-term and ethical performance. However, the length and complexity of ESG disclosures make them difficult to interpret and automate the analysis reliably. To support scalable and trustworthy analysis, this paper introduces ESG-Bench, a benchmark dataset for ESG report understanding and hallucination mitigation in large language models (LLMs). ESG-Bench contains human-annotated question–answer (QA) pairs grounded in real-world ESG report contexts, with fine-grained labels indicating whether model outputs are factually supported or hallucinated. Framing ESG report analysis as a QA task with verifiability constraints enables systematic evaluation of LLMs’ ability to extract and reason over ESG content and provides a new use case: mitigating hallucinations in socially sensitive, compliance-critical settings. We design task-specific Chain-of-Thought (CoT) prompting strategies and fine-tune multiple state-of-the-art LLMs on ESG-Bench using CoT-annotated rationales. Our experiments show that these CoT-based methods substantially outperform standard prompting and direct fine-tuning in reducing hallucinations, and that the gains transfer to existing QA benchmarks beyond the ESG domain.

\end{abstract}


\section{Introduction}

Accurate and trustworthy ESG (Environmental, Social, and Governance) reporting is increasingly essential for sustainable development, regulatory accountability, and ethical corporate conduct. ESG provides a framework for assessing how companies manage sustainability-related risks across environmental, social, and governance pillars \citep{de2023integration}. Once largely voluntary, ESG disclosure has become a legal requirement in many regions, most notably through EU regulations such as the Corporate Sustainability Reporting Directive and the Sustainable Finance Disclosure Regulation. This shift reflects growing expectations for transparency in corporate impacts on society and the environment \citep{niu2024government}. ESG reporting therefore plays a critical role in enabling compliance and supporting stakeholders’ evaluation of long-term performance \citep{arvidsson2022corporate, rossi2023independent}.

% Accurate and trustworthy ESG (Environmental, Social, and Governance) reporting is becoming essential for advancing sustainable development, regulatory accountability, and ethical corporate conduct.
% As shown in Figure \ref{fig:esg_pillars}, ESG is a framework used to assess how companies manage sustainability-related risks and opportunities across three key pillars: environmental, social, and governance \citep{de2023integration}.
% While ESG initially gained traction through voluntary disclosure and investor interest, it has increasingly become a legal and strategic requirement, particularly within the European Union. 
% Regulations such as the Corporate Sustainability Reporting Directive (CSRD) and the Sustainable Finance Disclosure Regulation (SFDR) now mandate ESG reporting. This shift reflects a broader societal push for transparency in how businesses affect social welfare, environmental sustainability, and ethical governance \citep{niu2024government}. In this context, ESG reporting has become a vital mechanism for aligning corporate behaviour with broader societal goals \citep{arvidsson2022corporate}. It enables companies to not only demonstrate compliance but also to build trust with stakeholders who increasingly demand responsible business conduct. In capital markets, where information asymmetry is a persistent issue, ESG disclosures help fill that gap by providing extra, non-financial information that complements traditional financial statements~\citep{rossi2023independent}.

Corporations now publish extensive ESG reports for investors, regulators, and the public \citep{assaf2024esg, seok2024esg}. However, the usefulness of these disclosures depends on their credibility and comparability. Third-party ESG rating agencies such as Sustainalytics and MSCI have been widely criticized for methodological opacity and inconsistency, with studies showing that their scores often diverge substantially even for the same company due to differences in indicator selection, weighting schemes, and data sources \citep{clementino2021companies, cort2020esg}. These controversies undermine stakeholder trust and highlight that ESG assessments are far from standardized. Combined with the growing length and complexity of sustainability reports, this inconsistency increases the need for scalable, transparent tools that can support reliable and evidence-grounded interpretation.

% In response to regulatory demands and rising stakeholder expectations, corporations now publish extensive ESG reports and establish dedicated sustainability teams. These reports serve a broad audience: investors use them to assess long-term risk and value \citep{assaf2024esg}, governments for compliance oversight, and the public to evaluate ethical and environmental accountability \citep{seok2024esg}. However, the effectiveness of ESG reporting depends not only on regulatory compliance but also on the credibility and comparability of the disclosed information. To aid interpretation, third-party agencies like Sustainalytics and MSCI offer ESG ratings, but their opaque methodologies have raised concerns about consistency and trustworthiness \citep{enwiki:1252485766, enwiki:1267063995}. Moreover, the use of different standards and rating methodologies across ESG evaluation systems often leads to inconsistent assessments of the same company, which weakens the signalling value of ESG disclosures \citep{clementino2021companies}. Such inconsistencies create difficulties for stakeholders attempting to make informed, comparative evaluations based on ESG information \citep{cort2020esg}. As ESG reports grow increasingly lengthy and complex, there is a growing demand for scalable and systematic tools to support efficient and consistent interpretation.

The emergence of large language models (LLMs) \citep{achiam2023gpt, dubey2024llama, dong2025safeguarding} offers new opportunities for automating the analysis of ESG disclosures at scale. However, the complexity and diversity of ESG reports pose significant challenges for reliable LLM deployment: (1) Companies may engage in \textit{greenwashing} \citep{yu2020greenwashing}, overstating their environmental initiatives to appear more sustainable, misleading investors and stakeholders about their true ESG impact. (2) ESG reports are \textit{rich in qualitative data} \citep{young2023looking}, requiring deep contextual understanding, industry-specific knowledge, and familiarity with regulatory frameworks, barriers that LLMs may struggle with due to their reliance on general knowledge. (3) ESG reports involve \textit{multi-modal processing} \citep{che2024predicting}, typically feature a mix of text, tables and graphics. (4) \textit{Long document retrieving and analysis} is also crucial \citep{ferjanvcivc2024textual}, as these documents often span hundreds of pages. 
LLMs remain limited in efficient document parsing, robust memory recall, and cross-sectional understanding in lengthy reports.

% With the advent of advanced technologies such as large language models (LLMs) \citep{achiam2023gpt, dubey2024llama, brown2020language}, new possibilities for enhancing the automatic evaluation of ESG reports have emerged. However, the complex and diverse nature of ESG reports introduces significant challenges for LLMs: 
\begin{figure}[t]
\centering
% \hfill
\includegraphics[width=0.38\textwidth]{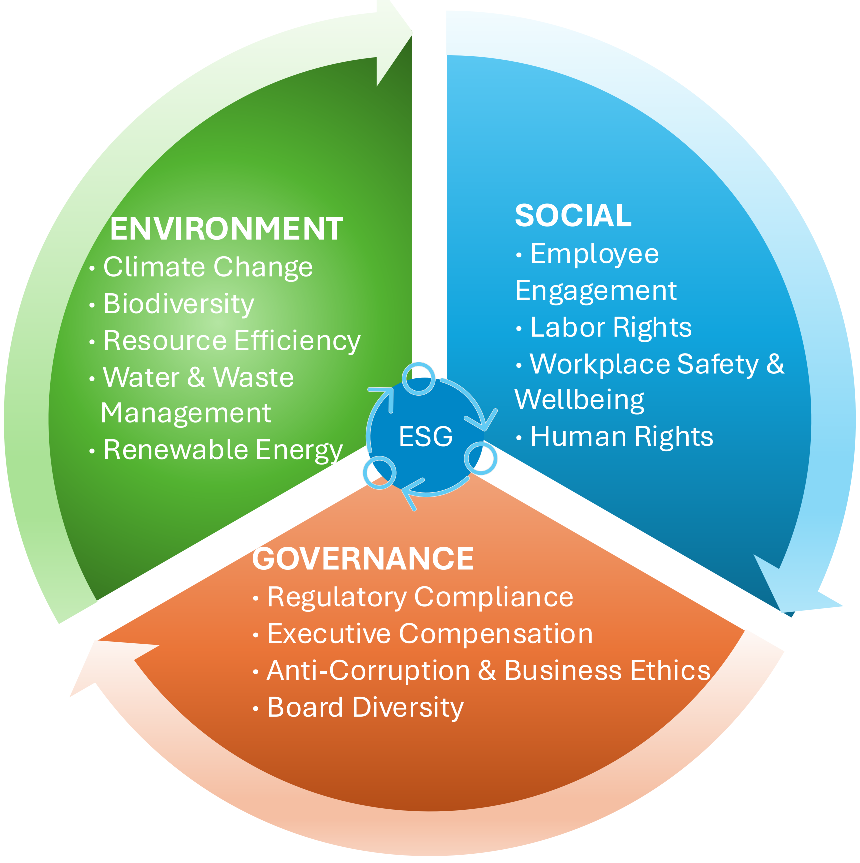}
% \hfill
  \caption{Core Pillars of ESG: Environmental, Social, and Governance Priorities. 
  }
  \label{fig:esg_pillars}
\end{figure}

% Despite the above domain-specific challenges for LLMs \citep{kamath2024llm}, these models also exhibit internal limitations. 

% Furthermore, LLMs face not only domain-specific challenges in ESG analysis but also internal limitations \citep{kamath2024llm}. One key issue is their reliance on parametric knowledge, referring to information encoded during pretraining, which may conflict with the factual content of ESG reports \citep{chen2024reckoning}. This misalignment can result in hallucinations, where generated responses are not grounded in the provided context.
% We define a hallucination as any model-generated answer that lacks support in the source document. Based on our annotation framework, we identify two types:
% (1) \textit{Additive hallucinations}, where the model introduces fabricated or unverifiable content;
% (2) \textit{Omissive hallucinations}, where the model fails to answer despite sufficient supporting context.
% While conceptually related to existing distinctions between factuality and faithfulness \citep{ji2023survey}, our framework formalizes these types through explicit human annotation for reliable benchmarking.

LLMs struggle with these demands due to limitations in document parsing, retrieval, and cross-sectional understanding, and also because they rely heavily on parametric knowledge that may conflict with the factual content of ESG reports \citep{kamath2024llm, chen2024reckoning}. This misalignment frequently leads to hallucinations, answers that are not grounded in the source document. 
We classify hallucinations into two types: (1) \textit{Additive hallucinations}, where the model introduces unsupported information, and (2) \textit{Omissive hallucinations}, where the model fails to answer despite relevant evidence. While related to notions of factuality and faithfulness \citep{ji2023survey}, these categories are formalized through explicit human annotation.

In this paper, we present ESG-Bench, a benchmark for hallucination-aware ESG question answering. We build the dataset through a model–then–annotator pipeline, establish a taxonomy of hallucination types, evaluate multiple LLMs on ESG-Bench, and propose a task-specific Chain-of-Thought (CoT) strategy for reducing hallucinations in long-context ESG analysis.
% In this paper, we introduce ESG-Bench, a benchmark for evaluating hallucination mitigation in ESG report question answering (QA). We construct the dataset through a model–then–annotator pipeline, define a clear taxonomy of hallucination types, and evaluate multiple LLMs on the benchmark. We further propose a task-specific Chain-of-Thought (CoT) prompting and fine-tuning strategy to reduce hallucinations in this high-stakes, long-context domain.
Our contributions are summarized below:
\begin{itemize}
\item \textbf{Benchmark Construction:} 
We present ESG-Bench, a benchmark dataset specifically designed for long-context QA and hallucination mitigation in ESG reporting. To the best of our knowledge, it is the first structured resource that supports both systematic evaluation and targeted mitigation of hallucinations in this socially and regulatory significant domain.

% We present ESG-Bench, a novel dataset specifically designed for context-based question-answering and hallucination detection tasks within the domain of ESG reporting. This effort marks the first attempt to create a structured dataset that facilitates both hallucination mitigation challenges and benchmarking in the ESG reporting context.

% \item \textbf{Task-Specific Strategy for Hallucination Mitigation:} We evaluate several representative LLMs on ESG-Bench for hallucination detection and compare their performance against human annotations. This provides a critical assessment of current model capabilities in identifying inaccuracies and fabrications within complex, real-world ESG documents.

% On our developed ESG-Bench, we evaluated several representative LLMs for hallucination detection. Additionally, we conducted comprehensive evaluations comparing the performance of these language models against human experts in detecting inaccuracies and fabrications within the dataset, providing a critical assessment of current inference capabilities versus human judgment.

\item \textbf{Task-Specific Strategy for Hallucination Mitigation:} We develop a fine-tuning approach based on task-specific CoT prompting and CoT-annotated reasoning traces. This method significantly improves factual grounding and reduces hallucinated outputs, demonstrating the effectiveness of structured reasoning in a domain-specific QA task.

\textbf{Empirical Evaluation:}
We fine-tune and evaluate multiple state-of-the-art LLMs on ESG-Bench, comparing their hallucination mitigation performance across both ESG and existing QA benchmarks. Our results highlight the unique challenges of long-context reasoning in ESG analysis and provide a robust assessment of model reliability in high-stakes compliance contexts.

\end{itemize}

\begin{figure*}[tp!]
\centering
\includegraphics[width=0.85\textwidth]{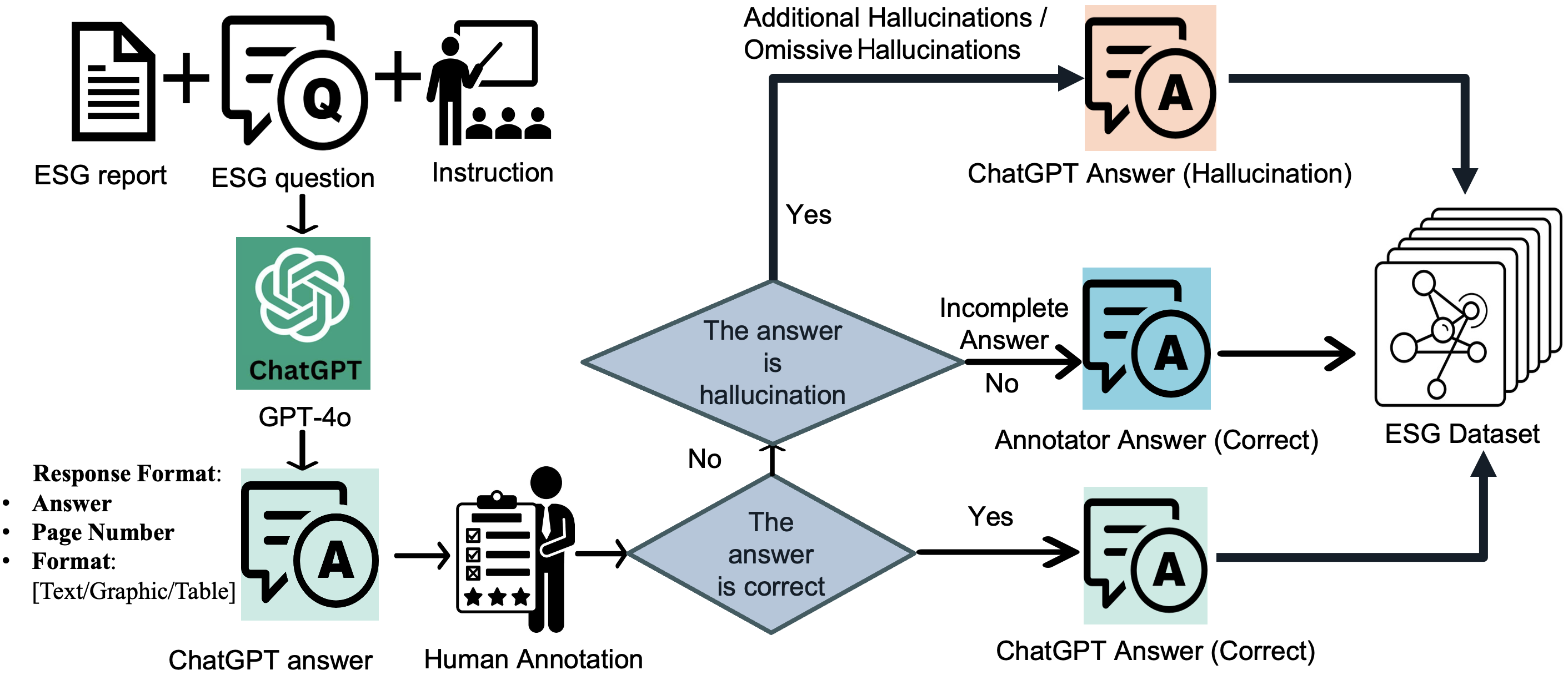}
  \caption{Workflow of ESG-Bench Construction. 
  % The process begins with feeding the GPT-4o with an ESG report, an ESG question, and an instruction. GPT-4o generates an answer, which is then reviewed by human annotators. If the answer is a hallucination, incomplete, or not found, annotators provide a corrected version. Otherwise, the correct answer is recorded. Both verified model-generated correct answers and human-corrected responses, along with their format (text, graphic, or table) and source pages, are stored in the ESG dataset for future use.
  }
  \label{fig1:flowchart}
\end{figure*}

\section{Related Work}

% Existing ESG-related work primarily  explores the use of Natural Language Processing (NLP) and LLMs to facilitate automated analysis and assessment of ESG reports. 
% ClimateQA utilized Transformer-based models as a question-answering system designed to extract relevant climate risk information from sustainability reports \citep{luccioni2020analyzing}. 
% Subsequently, another study incorporated OCR enhancements and NLP-based ranking models for financial decision-making  \cite{goel2020mining}.  
% Deep Search DocQA system allowed users to query ESG disclosures conversationally by integrating computer vision, NLP, and LLMs \citep{mishra2024esg}.
% CHATREPORT focused on integrating domain experts into the AI prompt engineering process while enabling automated sustainability report analysis. The benchmark aimed to achieve a high level of accuracy, with correctness and hallucination-free rates evaluated using human-annotated datasets. However, the aspect of hallucination detection remains unexplored \citep{ni2023chatreport}.
% ESG-Kor provided a benchmark dataset for Korean-language ESG research, filling the gap in non-English ESG datasets \citep{lee-etal-2024-esg}.
% Multilingual ESG issue identification was framed as a multi-label classification task to facilitate automated sustainability monitoring and responsible investment decision-making \citep{li-etal-2024-evaluating-multilingual}. 

\textbf{ESG-related Research:}
NLP and LLMs have been used to automate ESG disclosure analysis, including climate risk extraction \citep{luccioni2020analyzing}, financial QA \citep{goel2020mining}, conversational ESG querying \citep{mishra2024esg}, and automated sustainability report analysis such as ChatReport \citep{ni2023chatreport}. Recent work also explores multilingual and domain-specific ESG datasets \citep{lee2024esg, li2024evaluating}.
However, existing ESG QA resources focus on answer extraction and do not provide hallucination labels, CoT signals, or support long full-report contexts. ESG-Bench differs by offering human-verified hallucination annotations and tasks grounded in full corporate ESG documents.

% Existing ESG-related research leverages NLP, LLMs, and multimodal approaches to automate the analysis of sustainability reports. 
% Efforts include climate risk extraction \citep{luccioni2020analyzing}, financial decision-making via QA systems \citep{goel2020mining}, conversational querying of ESG disclosures \citep{mishra2024esg}, and automated sustainability report analysis \citep{ni2023chatreport}. These advances rely on Transformer models, OCR tools, multimodal integration, and expert-driven prompt design.
% Additionally, researchers have developed benchmarks and datasets to address language diversity in ESG research. For instance, recent studies explored Korean-language ESG analysis \citep{lee2024esg} and multilingual ESG issue classification \citep{li2024evaluating}, contributing to more inclusive and responsible investment decision-making across diverse linguistic contexts.
% While prior work has improved accuracy and reliability of automatic analysis via LLMs in ESG-related research, challenges related to hallucination mitigation remain largely unexplored. 
% In particular, there is no benchmark specifically designed to evaluate or mitigate hallucinations in long-form ESG reporting, where factual grounding and regulatory accuracy are critical.

% While advancements have improved accuracy and reliability of automatic analysis via LLMs in ESG-related research, challenges such as hallucination detection and the development of task-specific benchmarks remain largely unexplored.

\textbf{Hallucination Mitigation:}
Hallucination mitigation has been studied through calibration and self-evaluation \citep{kadavath2022language}, architectural interventions \citep{chrysostomou2024investigating}, entity-level verification \citep{zhao-etal-2020-reducing}, and uncertainty estimation \citep{xiao2020wat, farquhar2024detecting}. Complementary to these approaches, formal robustness verification has been applied to NLP models to provide certifiable guarantees on model behavior \citep{sun2023textverifier,wang2022deep, sun2024crowd}. Benchmarks such as HaluEval, TriviaQA, and BioASQ \citep{li2023halueval, joshi2017triviaqa, krithara2023bioasq} support evaluation across general domains but do not target long, heterogeneous ESG documents. ESG-Bench addresses this gap by enabling domain-specific assessment of factual grounding in ESG QA.

\textbf{Task-Specific CoT}
CoT prompting has been widely studied as a strategy for improving reasoning in LLMs \citep{wei2022chain}, with task-specific variants proposed for mathematical reasoning \citep{kojima2022large}, symbolic tasks \citep{zhang2023automatic}, fact verification \citep{lyu2023faithful}, and robustness considerations for LLM reasoning processes \citep{li2025pot, yi2024position}. However, existing approaches largely assume short or moderately sized contexts and do not address the evidence retrieval and verification demands of lengthy ESG reports. Our method introduces a CoT strategy tailored to long-context ESG disclosures, guiding models to extract and ground answers in the source text.

\section{ESG-Bench Construction }

In this section, we first describe each step of the construction process, with an overview of the pipeline illustrated in Figure~\ref{fig1:flowchart}. We then present a summary of the benchmark’s analysis and discuss its potential applications.

\subsection{ESG Report and Question Collection}

\subsubsection{(1) Report Collection:}

ESG reports were collected from {\textit{ResponsibilityReports.com}}\footnote{\url{https://www.responsibilityreports.com/}}, a publicly available online database of corporate sustainability disclosures. 
To ensure diversity, we selected reports from companies across various sectors, including finance, energy, technology, healthcare, consumer goods, and manufacturing etc. 
This industry diversity enables broad comparative analysis of ESG practices. 
For instance, companies in sectors with high environmental impacts, such as energy, mining, and manufacturing, often disclose more detailed environmental metrics, while those in finance and technology may focus more on governance and social initiatives.

\subsubsection{(2) Question Collection:}
\begin{figure}[hb!]
\centering
\includegraphics[width=0.47\textwidth]{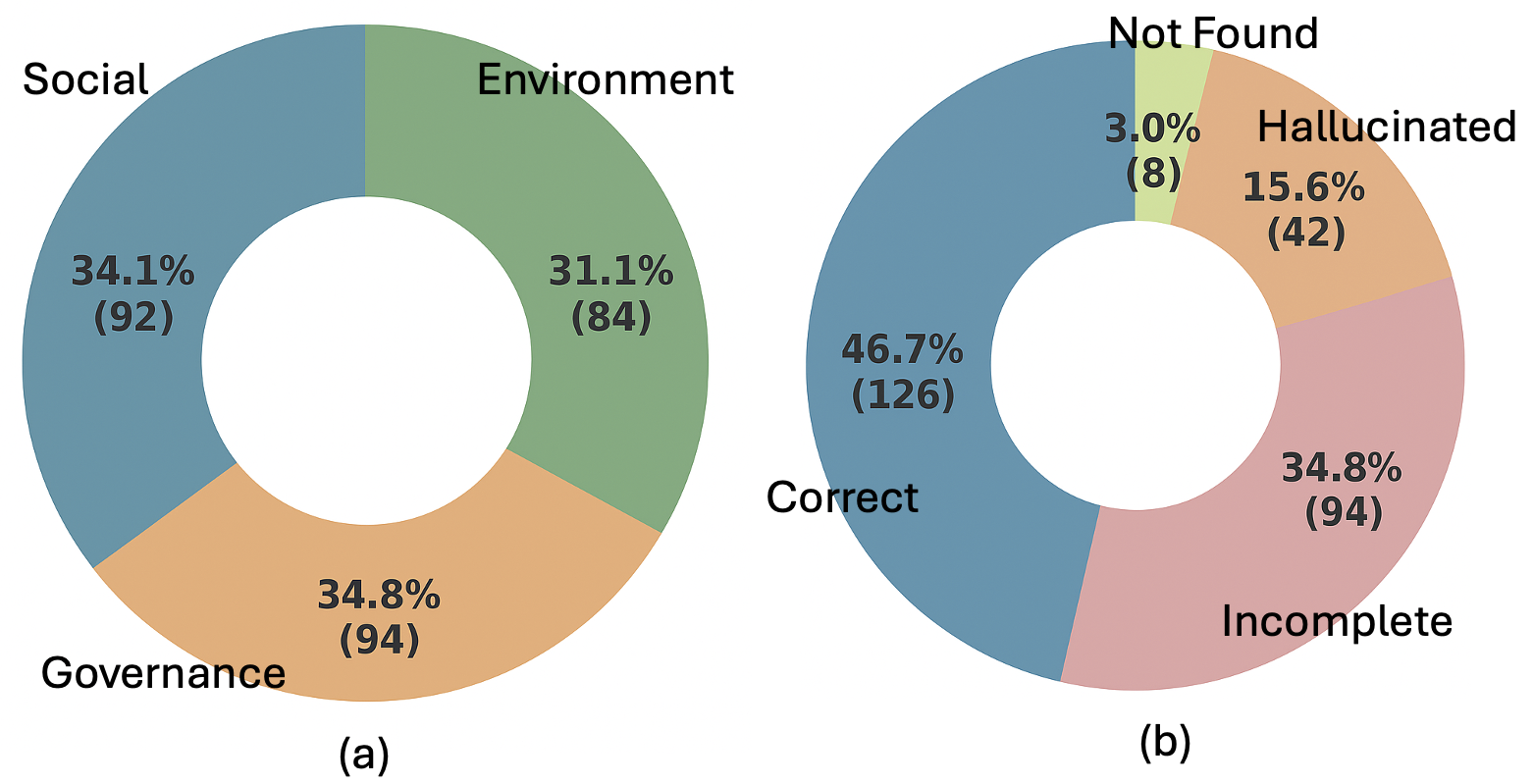}
  \caption{(a): Distribution of questions across ESG categories. (b): Distribution of QA pair labels.
  }
  \label{fig:question_distribution}
\end{figure}

The ESG-related questions are derived from multiple authoritative sources, including academic research \citep{mishra2024esg,parikh2024automatic, luccioni2020analyzing, arvidsson2022corporate, mishra2024statements, ni2023chatreport}, international non-profit organizations and corporate for  environmental reporting and risk management (\textit{Carbon Disclosure Project}\footnote{\url{https://www.cdp.net/en/disclose/question-bank}}, \textit{Caverion}\footnote{\url{https://www.caverion.com/contentassets/04e19da08cdf41a69b11ae2eb0a7832f/esg-questionnaire-final.pdf}}, \textit{Invest Europe}\footnote{\url{https://www.investeurope.eu/media/1777/invest-europe_esg_dd_questionnaire.pdf}}), and ChatGPT generated questions \citep{achiam2023gpt}. These sources ensure that the question set is aligned with real-world reporting practices and regulatory expectations. Questions are categorized into the three ESG pillars: Environmental, Social, and Governance, as shown in Figure~\ref{fig:esg_pillars}, to support structured coverage and domain-specific evaluation. Figure~\ref{fig:question_distribution}(a) presents the distribution of 270 questions across ESG categories.

\subsection{Model Instruction Design}
\label{subsection:gpt4o_qa}

Since ChatGPT-4o can process multi-modal reports, an instruction is designed to guide its response generation. The response is formatted with three key elements:
(1) \textit{Answer}: The generated response.
(2) \textit{Page Number}: The reference to the source page.
(3) \textit{Format}: Output type (Text, Graphic, or Table).
Generated responses are subsequently reviewed by human annotators for factual accuracy, contextual alignment, and formatting consistency (detailed in the next subsection).
The instruction design considers four key aspects: 

\subsubsection{(1) Question Diversity:}
A varied set of question templates was designed to evaluate hallucination behaviors across linguistic structures. These include open-ended forms (e.g., "How," "When," "Where") and directive prompts (e.g., "Break down," "Please describe"). Given the typical length and complexity of ESG reports, variation in question phrasing is intentionally used to increase the likelihood of hallucinations and assess the model’s sensitivity to questions. 

\subsubsection{(2) Domain-specific Selection:}
To account for variation across industries, ESG questions are categorized into general and sector-specific domains. A system prompt first elicits the company's core business description, which guides annotators in selecting the most relevant questions from a curated, domain-aligned question pool.

\subsubsection{(3) Expert Consultation:}
To ensure accuracy and contextual alignment, ESG domain experts manually reviewed the question pool. Their feedback informed the removal of redundant items, refinement of ambiguous phrasing, and alignment of questions with relevant regulatory frameworks. This expert input helped shape a question set that reflects best practices in ESG assessment and reporting.
% Manual review and feedback for question selection from ESG human experts ensures accuracy and contextual relevance. The curation process involves filtering duplicate or redundant questions to enhance uniqueness, aligning questions with regulatory frameworks to maintain compliance, ensuring clarity and specificity to minimize ambiguity. 
% % Details and examples of expert advice are demonstrated in Appendix \ref{appendix:expert_advice} and Table \ref{tab:esg_standardization}. 
% Their insights helped refine the framework and ensure that it aligns with best practices in ESG assessment. 

\subsubsection{(4) Iterative Refinement:}
The initial instruction framework underwent multiple rounds of revisions, incorporating expert feedback and real-world testing. The refinement process involved evaluating the interaction between ESG reports and model-generated responses, ensuring robustness and adaptability across different reporting scenarios.
% The initial instruction framework underwent multiple rounds of revisions, incorporating expert feedback and real-world testing. The refinement process involved evaluating the interaction between ESG reports and model-generated responses, ensuring robustness and adaptability across different reporting scenarios.
% By following the above phases, the designed instruction is provided in Appendix \ref{appendix:instuction_design}.

\begin{figure}[tp!]
\centering
  \includegraphics[width=1\columnwidth]{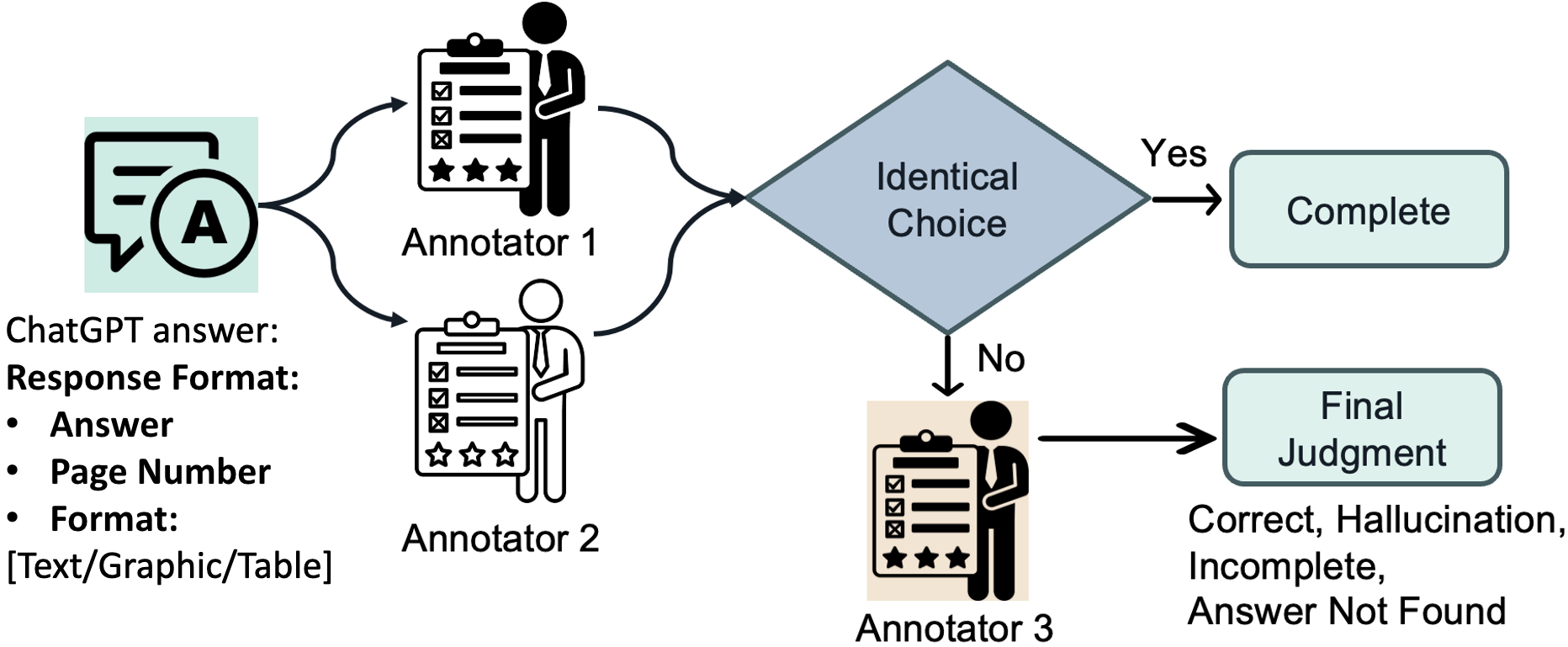}
  \caption{Annotator agreement.}
  \label{fig:annotator_agreement}
\end{figure}

\subsection{Human Annotation}
\label{subsection:human_annotation}

\subsubsection{(1) Annotator Recruitment}

To ensure high-quality annotations aligned with ESG standards, we recruited annotators with relevant expertise. These annotators were PhD-level students specializing in economics, sustainability, or related fields. Their expertise enabled them to accurately interpret and respond to questions based on the \textit{Global Reporting Initiative} (GRI) standards \citep{hedberg2003global}, a widely used framework for sustainability reporting. 

The recruitment prioritized individuals with:
(a) Proficiency in ESG concepts and reporting structures;
(b) Experience with financial and non-financial disclosures;
(c) Prior academic or professional engagement with GRI-based reporting.
This selection ensured consistent and accurate evaluation across diverse industry disclosures.
% By selecting well-qualified annotators, we aimed to enhance the accuracy and reliability of the question-answering process in ESG-related annotation.

\subsubsection{(2) Annotation Procedure}

The annotation process followed a structured workflow, as illustrated in Figure \ref{fig:annotator_agreement}. Each model-generated response, including the proposed answer, cited page number, and content format (text, table, or graphic), was independently reviewed by two annotators using a predefined evaluation criteria.
Each response was assigned one of the following labels:
(a) \textit{Correct:} Fully supported by context.
(b) \textit{Hallucination:} The response contains information that is fabricated or not supported by the source.
(c) \textit{Incomplete:} Partially accurate but missing key information.
(d) \textit{Answer Not Found:} The model returned "Not provided" despite a valid source answer.

% To ensure the reliability of our hallucination detection benchmark, we establish a structured annotation process involving multiple annotators. Each generated response from the model is evaluated using a \textit{three-stage annotation} framework, as illustrated in Figure \ref{fig:annotator_agreement}:

% (a) \textbf{\textit{Initial annotation by two annotators:}} Each response is independently assessed by two annotators based on predefined evaluation criteria, including factual correctness, completeness, and alignment with the reference source in the report. Annotators evaluate the answer, page number, and format (Text/Graphic/Table) to ensure consistency in evaluation. 
% % The detailed content of annotator guidelines can be referred to in Appendix \ref{appendix:annotator_guidelines}.
% The response can be categorized into one of four possible labels:
% (i) \textit{Correct:} The response is factually accurate and aligns with the reference.
% (ii) \textit{Hallucination:} The response contains information that is either factually incorrect or not supported by the background knowledge.
% (iii) \textit{Incomplete:} The response is partially correct but lacks necessary details.
% (iv) \textit{Answer Not Found:} The model response is "Not provided" but the answer can be found in the background knowledge.

\subsubsection{(3) Conflict Resolution}
When both annotators agreed on the assigned label, the annotation was finalized. In cases of disagreement, a third annotator resolved the conflict through majority voting. For all instances labeled as \textit{Hallucination}, \textit{Incomplete}, or \textit{Answer Not Found}, a corrected answer was written by the annotator, following the protocol in Figure~\ref{fig1:flowchart}.

\subsubsection{(4) Inter-Annotator Agreement}
\begin{figure}[tp!]
\centering
  \includegraphics[width=0.9\columnwidth]{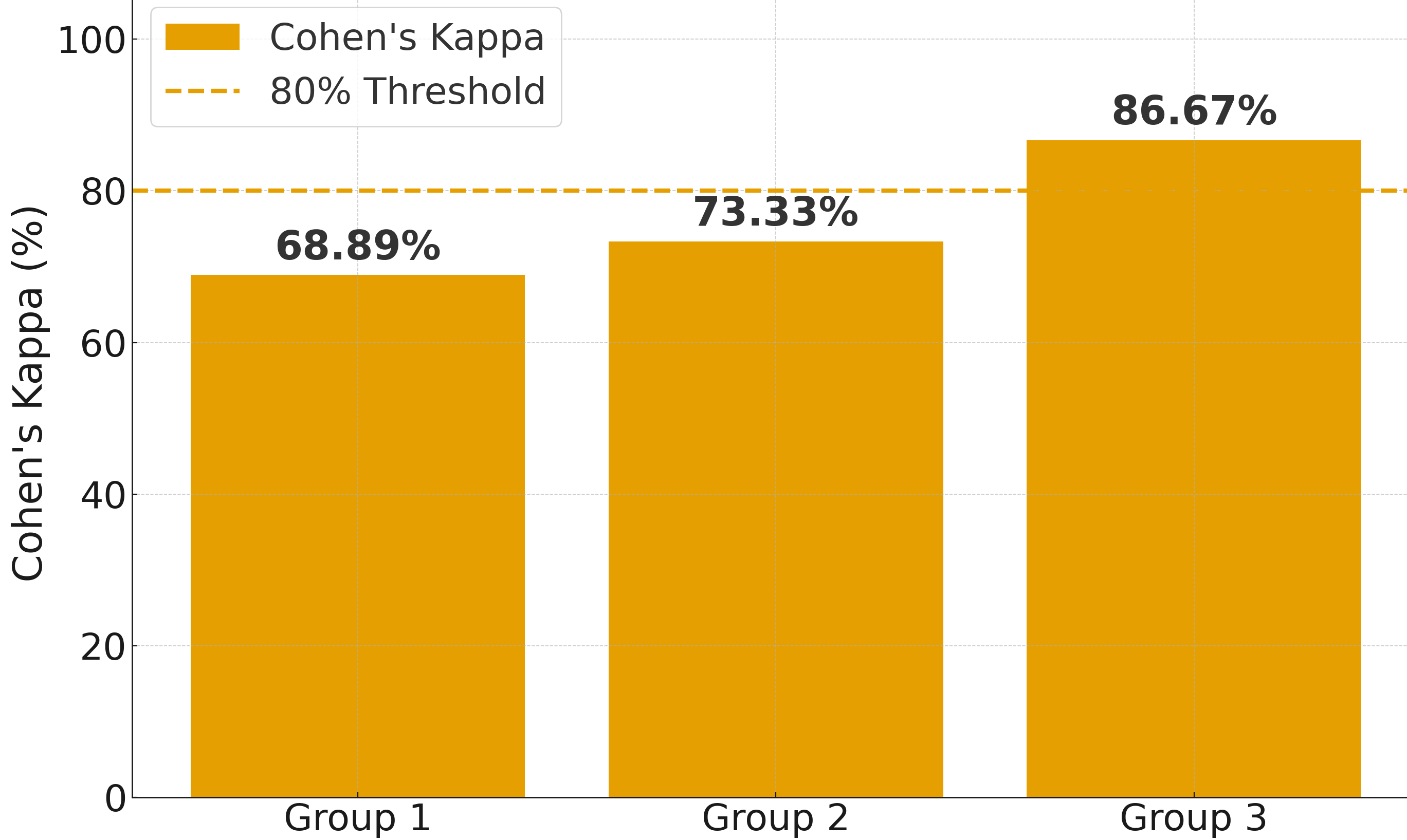}
  \caption{Cohen's Kappa across 3 annotator groups.}
  \label{fig:cohen_kappa}
\end{figure}

Cohen’s Kappa \citep{cohen1960coefficient} was adopted as the statistical measure to assess inter-annotator agreement, accounting for the possibility of chance agreement. This analysis evaluated consistency across three annotator groups, with the results presented in Figure~\ref{fig:cohen_kappa}. Group 3 achieved near-perfect agreement (86.67\%), while Groups 1 and 2 showed substantial agreement (68.89\% and 73.33\%, respectively). These scores confirm strong alignment among annotators and demonstrate the reliability and robustness of the annotation process.
% \subsubsection{(b) Agreement check:}
%  If both annotators make an identical judgment, the annotation process is considered complete, and the response is assigned its final labels. 

% \subsubsection{(c) Resolution by a third annotator:}
% If the first two annotators disagree, a third annotator is introduced to provide a final judgment via majority voting. 
% Notably, if the label is "correct", the correct answer is ChatGPT generated answer. Otherwise, if the label is "Hallucination", "Incomplete" or "Answer Not Found", the human annotator would provide a corrected answer as shown in Figure \ref{fig1:flowchart}. 

\subsection{Benchmark Analysis and Usage}

ESG-Bench is a high-quality resource for evaluating ESG-related QA systems, with an emphasis on answer correctness, completeness, and hallucination mitigation. It comprises two complementary versions, each designed to support different evaluation objectives.

\subsubsection{(1) Report-based Dataset}
This version contains 270 QA instances drawn from 94 unique ESG reports published between 2020 and 2024 (with some reports reused across questions). Each instance includes a question grounded in the report, a response generated by ChatGPT-4o, and a human assessment. Annotators verify the factual accuracy of the answer, identify the supporting source pages, and note the format in which the relevant information appears (e.g., text, table, or graphic).
In Figure \ref{fig:question_distribution} (b), the distribution of QA pair labels shows that 228 responses (84.44\%) were either correct as generated or made correct through human annotation, while 42 (15.56\%) were classified as hallucinations. Of all annotated responses, 46.7\% were deemed correct, 34.8\% incomplete, 3.0\% had an answer not found (omissive hallucinations), and 15.6\% were factually hallucinated.

\subsubsection{(2) Dataset for Hallucination Mitigation Task}
This version of ESG-Bench is designed to support the evaluation of hallucination mitigation in LLMs. Each example consists of a background passage, an ESG-related question, and a model-generated answer. Human annotators evaluate whether the response is grounded in the provided context and categorize hallucinations into two types: factually incorrect answers and answers that are unsupported by the background knowledge.

\begin{figure}[tp!]
\centering
  \includegraphics[width=0.9\columnwidth]{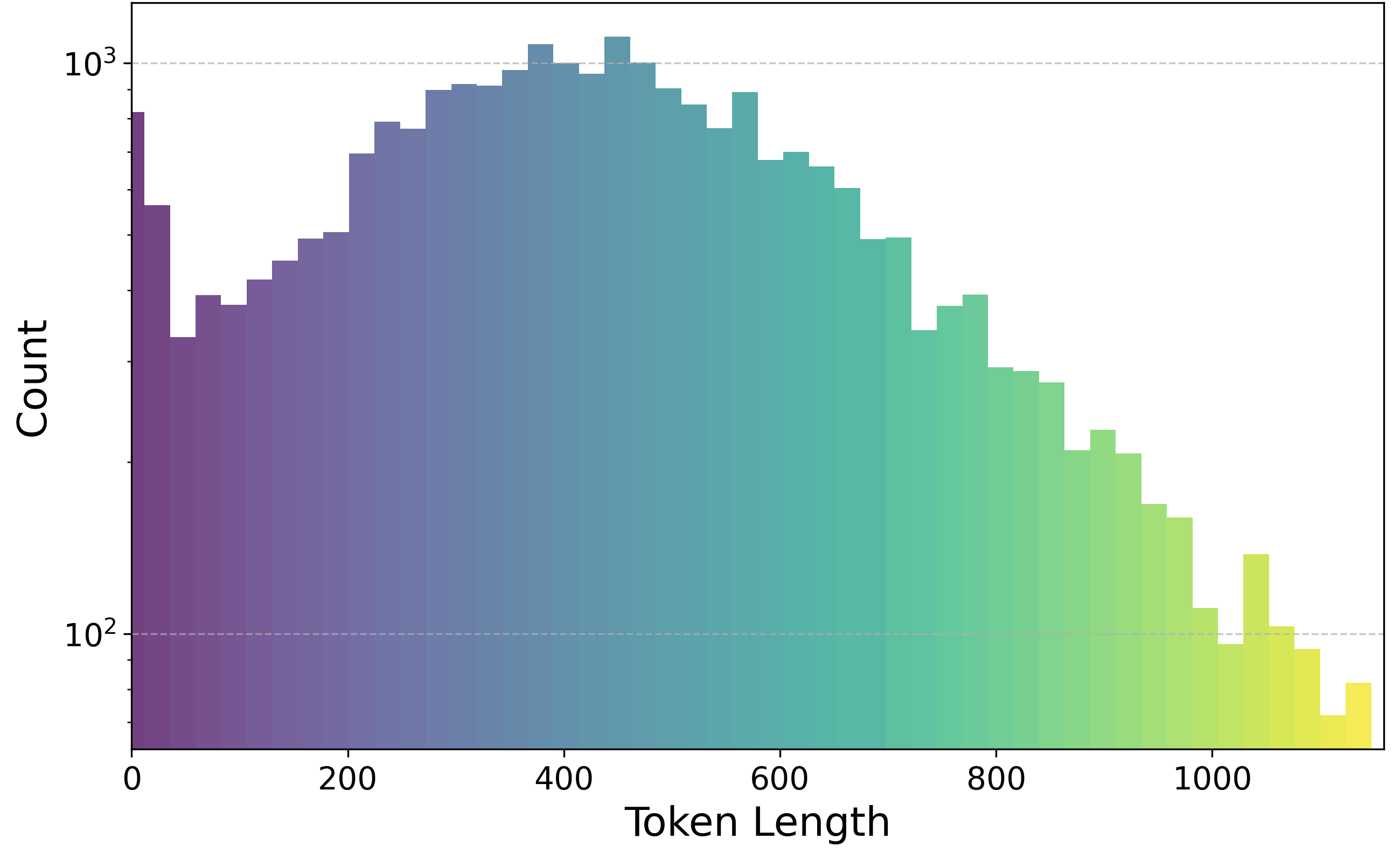}
  \caption{Length distribution of context knowledge.}
  \label{fig:length_distributions}
\end{figure}

As shown in Figure \ref{fig:length_distributions}, the background passages vary in length, with a maximum of 46,562 tokens and an average of 2,604 tokens. Answer lengths range from 3 to 3,362 tokens, with a mean of approximately 614 tokens. For readability, the figure visualises distributions after applying an IQR-based outlier filter, which removes 2.39\% of extreme-length instances; all reported statistics are computed on the full dataset. In total, the dataset includes 1,358 correct responses and 25,516 hallucinated responses. Among the hallucinated instances, 21,724 were labeled as unsupported by the given context, while 3,706 were identified as factually incorrect.

\subsubsection{(3) ESG-Bench Usage}

ESG-Bench supports both research and practical applications. Annotator-corrected responses enable fine-tuning of ESG-specific QA models for improved factual grounding, while hallucination labels aid in developing mitigation strategies. The dataset also serves as a benchmarking tool for evaluating answer accuracy, retrieval robustness, and format-specific performance. Its hallucination-focused variant supports training classifiers to detect unsupported or incorrect content. In practice, ESG-Bench can assist in corporate ESG audits and compliance verification, and it provides a valuable resource for training summarization models on long ESG documents.

% ESG-Bench supports a range of research and practical applications. First, the inclusion of annotator-corrected responses enables the training and fine-tuning of \textit{ESG-specific QA models} for improved factual accuracy and contextual grounding. The hallucination labels facilitate the development of dedicated mitigation techniques. Second, the dataset can serve as a \textit{benchmarking tool} to evaluate information extraction systems across multiple dimensions, such as answer correctness, retrieval robustness, and format-specific performance, with source page numbers and answer formats providing added granularity. Third, the hallucination-focused version of the dataset allows for training and evaluating classifiers that \textit{distinguish factual from hallucinated content}, with support for different hallucination types (e.g., unsupported or factually incorrect). Beyond model development, ESG-Bench has potential real-world applications in \textit{corporate ESG audits and compliance verification}, helping stakeholders assess AI-generated disclosures against sustainability standards. Lastly, it offers a valuable resource for training \textit{summarization models} that extract structured or unstructured ESG insights from long documents.

% \begin{figure}[t!]
%     \centering
%     \subfloat{%
%         \includegraphics[width=0.48\textwidth]{figures/knowledge_lengths_filtered.png}
%     }
%     \hfill
%     \subfloat{%
%         \includegraphics[width=0.48\textwidth]{figures/hallucinated_lengths.png}
%     }
%     \caption{\textit{ESG-Bench: Length distribution of context knowledge and hallucinated answers.}}
%     \label{fig:length_distributions}
% \end{figure}

\section{Strategies Toward Hallucination Mitigation}

% Our objective is to reduce hallucinations in ESG QA by generating answers that are strictly grounded in a given context. In our setup, each question is paired with an ESG-related passage, and the correct answer is either (1) explicitly supported by the context, or (2) labeled “Not provided” when the required information is absent. 
% We define hallucination as any failure of the model to correctly ground its answer in the provided document. This includes two failure modes: (1) \textit{hallucinated additions}: the model generates content not found in the context, and (2) \textit{hallucinated omission}:the model outputs “not provided” even though the context does contain sufficient information. Both cases reflect a breakdown in the model's alignment with the source knowledge.
% To mitigate hallucination, we propose a progressively deepening methodology consisting of three stages: supervised fine-tuning with grounded answers, CoT prompting for structured inference, and CoT-based fine-tuning on annotated reasoning traces. Each stage builds upon the previous one, improving the model’s ability to reason over and align with the context while reducing unsupported outputs.

Our goal is to reduce hallucinations in ESG QA by ensuring answers are strictly grounded in the given context. Each question is paired with an ESG-related passage, and the correct answer is either supported by the text or labeled “Not provided” if absent. 
% Hallucinations occur when models generate unsupported content (hallucinated additions) or fail to recognize existing information (hallucinated omissions). 
We propose a three-stage approach: supervised fine-tuning, CoT prompting, and CoT-based fine-tuning. 
% This progressively enhances contextual alignment and reduces hallucinated outputs.

\subsection{Phase 1: Supervised Fine-tuning with Contextual Grounding}

We begin by fine-tuning a LLM on our ESG QA dataset, where each instance consists of a report context, a question, and a human-annotated answer. The answer is either explicitly supported by the passage or labeled “Not provided” if the relevant information is missing. Our approach builds on techniques from factual QA fine-tuning \citep{tian2023fine}.
This training encourages the model to attend to explicit textual evidence, learning to generate grounded answers when possible and to abstain otherwise. We frame this as a standard sequence-to-sequence task, optimizing likelihood over the ground-truth answers. 
Although this baseline reduces hallucinations compared to zero-shot models (see Experimental Results), it can still produce overconfident outputs when evidence is ambiguous or incomplete, motivating further refinement.

% This baseline helps reduce hallucinations compared to zero-shot models as shown in Experimental Results, but we observe limitations, particularly when the model responds over-confidently to ambiguous or partial cues. These limitations motivate our next step.

% \subsection{Phase 2: CoT Prompting for Structured Inference}
\subsection{Phase 2: CoT Prompting and Fine-tuning}

% To enhance contextual reasoning, we introduce CoT prompting at inference time. Rather than instructing the model to answer directly, we guide it through intermediate reasoning steps to evaluate the sufficiency of evidence in the passage.
% Inspired by prior work showing that CoT can improve factual and multi-step reasoning \citep{kojima2022large, wei2022chain}, we design two task-specific CoT prompting templates tailored to the ESG domain:
To improve the model’s reasoning and evidence assessment, we introduce CoT prompting at inference time. Instead of producing an answer directly, the model is guided by intermediate steps to evaluate whether the passage contains sufficient information. Based on prior work showing the benefits of CoT for factual reasoning \citep{kojima2022large, wei2022chain}, we design two ESG-specific prompting formats:

(1) \textit{Two-step CoT:}
\begin{quote}
1. Determine if the report provides an answer to the question: \texttt{\{answerable\}} \\
2. Based on your reasoning, the correct answer should be: \texttt{\{answer\}}
\end{quote}
(2) \textit{Four-step CoT:}
\begin{quote}
1. Identify the key topic or entity mentioned in the question: \texttt{\{topic\}} \\
2. Search the report for sentences or paragraphs relevant to that topic: \texttt{\{report summary\}} \\
3. Determine if the report provides an answer to the question: \texttt{\{answerable\}} \\
4. Based on your reasoning, the correct answer should be: \texttt{\{answer\}}
\end{quote}
In both formats, the \texttt{\{answerable\}} label and the final \texttt{\{answer\}} are treated as ground-truth annotations provided by human annotators, while the topic in the four-step template is generated automatically by GPT-4o.
Unlike general CoT approaches that generate free-form reasoning paths \citep{wang2023self, huang2023towards}, our templates are explicitly tailored to ESG contexts. This structure enhances factual consistency and reasoning interpretation.

% Unlike prior CoT approaches that rely on LLMs to generate reasoning paths automatically \citep{wang2023self, huang2023towards}, our prompts are explicitly designed for the ESG domain and structured to address the specific challenges of document grounding and abstention. This domain-aware design makes the reasoning process more controllable and interpretable, while improving factual alignment.

% \subsection{Phase 3: Fine-tuning on CoT-Annotated Reasoning Traces}

To further strengthen internal consistency, we fine-tune the model on a curated subset of QA pairs annotated with explicit chain-of-thought (CoT) rationales. Each example includes intermediate reasoning steps leading to either a grounded answer or a justified “Not provided” conclusion. This step builds on recent work showing that CoT supervision improves model consistency and factual accuracy \citep{chung2024scaling}.
By learning from explicit human reasoning paths, the model is encouraged to internalize structured decision-making rather than relying on surface-level patterns or implicit heuristics. This reinforces contextual alignment, particularly when evidence is indirect or sparse.

Each stage of our methodology addresses specific limitations of the previous one. Supervised fine-tuning instills grounding ability but does not make the reasoning process transparent. CoT prompting introduces structured inference and improves contextual deliberation, but remains dependent on external prompting during inference. CoT-based fine-tuning internalizes these reasoning structures, enhancing robustness, interpretability, and reducing both hallucinated additions and omissions.
This staged framework constitutes a progressively refined strategy for hallucination mitigation in document-grounded ESG QA task.

\section{Experiments}

\subsection{Experimental Setting}

Our code is publicly available at \url{https://github.com/GateNLP/ESG_Bench}.

\subsubsection{Evaluation Models}

We evaluate several state-of-the-art LLMs using the ESG-Bench benchmark. Specifically, we assess three prominent models: Llama-3.2-3B Instruct \citep{dubey2024llama}, Gemma-2-2B-it \citep{team2024gemma}, and Mistral-7B-Instruct-v0.3 \citep{jiang2023mistral}. These models are tested on their ability to generate responses while identifying hallucinations by evaluating uncertainty. 
% Additionally, we leverage GPT-4o as a comparative baseline, using it to compare the generated responses against reference answers in sentence-length generation tasks. 

% We execute the generation process of correct and hallucinated samples using OpenAI ChatGPT-4o. We use a temperature of 1.0 to generate samples and set the maximum number of tokens for generation to 256. Moreover, we set the frequency penalty to zero and top-p to 1.0. For evaluation, we set the temperature to zero for all models to reduce output randomness and ensure more focused and deterministic outputs. In the following, we first conduct hallucination recognition experiments, then propose several potentially useful strategies to improve the recognition, and finally we perform qualitative analysis to understand the hallucination in LLMs.

\subsubsection{Datasets}

To assess the generalizability of our hallucination mitigation approach beyond the ESG domain, we incorporate additional datasets following the evaluation setup in \citet{farquhar2024detecting}. 
% We sample 400 instances from the test or validation split of each dataset. 
The selected benchmarks include: (1) \textbf{BioASQ} \citep{krithara2023bioasq}, a biomedical QA dataset focused on scientific literature from the life sciences; and (2) \textbf{HaluEval} \citep{li2023halueval}, a benchmark specifically designed to assess hallucination in LLM outputs across diverse tasks and domains. 
As shown in Table \ref{tab:dataset_stats}, we finetune the models on each dataset's respective training set and evaluate on a test set. Each test set is approximately balanced between WA (With Answer) and WoA (Without Answer) instances, allowing us to assess the model’s ability not only to generate accurate answers but also to abstain appropriately when sufficient information is unavailable. Notably, for ESG-Bench, we split the data by reports to prevent data leakage between training and test sets, as individual reports may contain overlapping content that could compromise evaluation integrity.

\begin{table}[t!]
\centering
    \renewcommand{\arraystretch}{1.2}
    \resizebox{0.45\textwidth}{!}{
\begin{tabular}{lccccc}
\hline
\textbf{Dataset} & \textbf{Train} & \textbf{Test} & \textbf{WA in Test} & \textbf{WoA in Test} \\
\hline
Halueval   & 16,000 & 400 & 196 & 204 \\
Bioasq     & 6,040  & 400 & 198 & 202 \\
ESG-Bench  & 2,807  & 300 & 142 & 158 \\
\hline
\end{tabular}}
\caption{Dataset statistics: number of training examples, test examples, and distribution of WA (With Answer) and WoA (Without Answer) in the test sets.}
\label{tab:dataset_stats}
\vspace{-5mm}
\end{table}

\subsubsection{Implementation Details}

All our experiments are conducted on high-performance computing clusters equipped with NVIDIA\textsuperscript{\textregistered} GH200 480GB GPUs and ARM\textsuperscript{\textregistered} Neoverse-V2 CPUs (72 cores, 3.41\,GHz). We fine-tune LLMs using the HuggingFace \texttt{transformers} and \texttt{trl} libraries. We use the AdamW optimizer, a learning rate of 2e-5, and a warmup ratio of 0.1. Each model is fine-tuned for 20 epochs with a batch size of 32, keeping all other hyperparameters consistent with the pretraining stage. 
For Chain-of-Thought (CoT) training, data are generated using greedy decoding to ensure reproducibility.
% We apply LoRA with rank 16 and dropout 0.2 to selected transformer modules (e.g., \texttt{q\_proj}, \texttt{k\_proj}, \texttt{v\_proj}, \texttt{o\_proj}, etc.), and enable gradient checkpointing to reduce memory overhead. For each dataset, 10\% of training examples are held out for validation. All prompts follow a CoT-style structure and are tokenized using chat templates consistent with the base model family (e.g., Mistral, LLaMA, or Gemma).

% We pretrain our model for 20 epochs using the AdamW optimizer with a batch size of 256, a learning rate of 0.001, weight decay of $10^{-5}$, and a one-cycle learning rate scheduler. Subsequently, we fine-tune the model for 500 epochs on downstream tasks using a batch size of 32, keeping all other hyperparameters consistent with the pretraining stage.

\subsubsection{Evaluation Metrics}

% \textbf{Hallucination uncertainty metrics: }
% (1) \textbf{Area Under the Receiver Operating Characteristic Curve (AUROC)} \citep{honovich2022true}: Measures the trade-off between true and false positive rates in hallucination detection. Higher values indicate better discrimination.
% (2) \textbf{P(False):} Represents the probability that a generated response is factually incorrect or hallucinated, reflecting the model’s confidence in incorrect answers.
% (3) \textbf{Regular Entropy:} Quantifies uncertainty based on the probability distribution of possible responses. 
% (4) \textbf{Semantic Entropy:} Extends regular entropy by measuring uncertainty in semantic embeddings rather than token-level probabilities.
% (5) \textbf{Cluster Entropy: }Evaluates how well generated outputs align with semantic clusters. Higher entropy suggests greater dispersion across meaningfully distinct clusters, which may indicate hallucination.
% (6) \textbf{Eigenscore:} \citep{chen2024inside} Assesses semantic consistency by analyzing the covariance of sentence embeddings across multiple generated responses. Higher variation suggests greater semantic divergence.

% \textbf{Hallucination detection metrics:}
(1) \textit{WA Accuracy:} The proportion of correct predictions for instances where an answer exists in the context. This measures the model's ability to provide faithful, grounded responses.
(2) \textit{WoA Accuracy:} The proportion of correct predictions for instances where no answer is available in the document ("Not provided."). This reflects the model’s ability to abstain and avoid hallucination when insufficient information is present.
(3) \textit{Balanced Accuracy:} The average of WA and WoA accuracy, this metric reflects the model’s overall ability to perform well on both answerable and unanswerable cases.
(4) \textit{F1 Score:} This score captures the tradeoff between precision (avoiding false alarms) and recall (catching actual hallucinations).

% (1) \textbf{Accuracy: }The overall proportion of correct predictions (both hallucinated and faithful answers).
% (2) \textbf{Precision:} The proportion of outputs predicted as hallucinated that are actually unsupported by the document. A high precision means the system rarely mislabels faithful answers as hallucinated (low false positives).
% (3) \textbf{Recall:} The proportion of all hallucinated (unsupported) answers that are correctly identified. A high recall indicates the system successfully catches most hallucinated answers (low false negatives).
% (4) \textbf{F1 Score:} The harmonic mean of precision and recall, balancing both aspects of mitigation.

\begin{table*}[t!]
\centering
\renewcommand{\arraystretch}{1.3}
\setlength{\tabcolsep}{3pt}
\resizebox{1.0\textwidth}{!}{
\begin{tabular}{ll|cccccccc|cccccccc|cccccccc}
\toprule
\textbf{Dataset} & \textbf{Setting} 
& \multicolumn{8}{c}{\textbf{LLaMA}} 
& \multicolumn{8}{c}{\textbf{Gemma}} 
& \multicolumn{8}{c}{\textbf{Mistral}} \\
 &  & \multicolumn{2}{c}{\textbf{True}}
    & \multicolumn{2}{c}{\textbf{False}}
    & \multicolumn{3}{c}{\textbf{Acc (\%)}}
    % & \textbf{ACC (\%)} 
    & \textbf{F1 (\%)}
    & \multicolumn{2}{c}{\textbf{True}}
    & \multicolumn{2}{c}{\textbf{False}}
    & \multicolumn{3}{c}{\textbf{Acc (\%)}}
    % & \textbf{ACC (\%)} 
    & \textbf{F1 (\%)}
    & \multicolumn{2}{c}{\textbf{True}}
    & \multicolumn{2}{c}{\textbf{False}}
    & \multicolumn{3}{c}{\textbf{Acc (\%)}}
    % & \textbf{ACC (\%)} 
    & \textbf{F1 (\%)} \\
 &  & WA & WoA & WA & WoA & WA & WoA & OA & OA
    & WA & WoA & WA & WoA & WA & WoA & OA & OA
    & WA & WoA & WA & WoA & WA & WoA & OA & OA \\
\cmidrule(lr){3-10} \cmidrule(lr){11-18} \cmidrule(lr){19-26}
\midrule
\multirow{4}{*}{ESG-Bench} 
& WoF             
& 96  & 132 & 46 & 26 & 67.61 & 83.54 & 76.00 & 65.23
& 83  & 130 & 59 & 28 & 58.45 & 82.28 & 71.00 & 60.84
& 110 & 132 & 32 & 26 & 77.46 & 83.54 & 80.67 & 69.64 \\
& SFT             
& 115 & 157 & 27 & 1  & 80.99 & 99 & 90.67 & 73.68
& 63  & 127 & 79 & 31 & 44.36 & 80.38 & 63.33 & 53.87
& 87  & 153 & 55 & 5  & 61.27 & 96.83 & 80.00 & 64.58 \\
& CoT (2)   
& 119 & 158 & 23 & 0  & 83.80 & \textbf{100} & 92.33 & 75.01
& 108 & 110 & 34 & 48 & 76.05 & 69.62 & 72.67 & 66.42
& 101 & 157 & 41 & 1  & 71.12 & \textbf{99.37} & 86.00 & 69.35 \\
& CoT (4)   
& 131 & 157 & 11 & 1  & \textbf{92.52} & 99.37 & \textbf{96.00} & \textbf{78.62}
& 129 & 147 & 13 & 11 & \textbf{90.85} & \textbf{93.04} & \textbf{92.00} & \textbf{77.09}
& 115 & 155 & 27 & 3  & \textbf{80.99} & 98.10 & \textbf{90.00} & \textbf{73.50} \\
\midrule
\multirow{4}{*}{HaluEval} 
& WoF             
& 159 & 128 & 37 & 76 & 81.12 & 62.75 & 71.75 & 67.61
& 171 & 189 & 25 & 15 & 87.24 & 92.64 & 90.00 & 75.97
& 177 & 164 & 19 & 40 & 90.30 & 80.39 & 85.25 & 75.34 \\
& SFT             
& 167 & 140 & 29 & 64 & 85.20 & 68.63 & 76.75 & 70.73
& 183 & 85  & 13 & 119& 93.37 & 41.67 & 67.00 & 67.59
& 183 & 187 & 13 & 17 & 93.27 & 91.67 & 92.50 & 78.62 \\
& CoT (2)   
& 178 & 136 & 18 & 68 & \textbf{90.82} & 66.67 & 82.67 & 72.89
& 187 & 90  & 9  & 114& 95.41 & 44.11 & 69.25 & 69.24
& 187 & 183 & 9  & 21 & 95.41 & 89.70 & 92.50 & 79.25 \\
& CoT (4)   
& 170 & 162 & 26 & 42 & 86.73 & \textbf{79.41} & \textbf{83.00} & \textbf{73.52}
& 187 & 189 & 9  & 15 & \textbf{95.41} & \textbf{92.64} & \textbf{94.00} & \textbf{79.71}
& 188 & 202 & 8  & 2  & \textbf{95.91} & \textbf{99.02} & \textbf{97.50} & \textbf{80.87} \\
\midrule
\multirow{4}{*}{BioASQ} 
& WoF             
& 152 & 177 & 46 & 25 & 76.77 & 87.62 & 82.25 & 83.29
& 159 & 200 & 39 & 2  & 80.30 & \textbf{99.01} & 89.75 & 73.90
& 177 & 197 & 21 & 5  & 89.39 & 97.52 & 93.50 & 77.85 \\
& SFT             
& 164 & 161 & 34 & 41 & 82.82 & 79.70 & 81.25 & 71.91
& 149 & 149 & 49 & 53 & 75.25 & 73.76 & 74.50 & 67.29
& 180 & 202 & 18 & 0  & 90.90 & \textbf{100} & 95.50 & 78.90 \\
& CoT (2)   
& 123 & 202 & 75 & 0  & 62.12 & \textbf{99.50} & 81.00 & 65.69
& 190 & 151 & 8  & 51 & 95.96 & 74.75 & 85.25 & 77.08
& 191 & 201 & 7  & 1  & 96.46 & 99.50 & 98.00 & 81.40 \\
& CoT (4)   
& 176 & 202 & 22 & 0  & \textbf{88.89} & \textbf{99.50} & \textbf{94.25} & \textbf{77.97}
& 198 & 200 & 0  & 2  & \textbf{100} & \textbf{99.01} & \textbf{99.50} & \textbf{82.97}
& 192 & 201 & 6  & 1  & \textbf{96.96} & 99.50 & \textbf{98.25} & \textbf{81.63} \\
\bottomrule
\end{tabular}}
\caption{Full performance comparison across 3 LLMs under different finetuning strategies. WoF: Without Finetuning, SFT: Supervised finetuning, CoT ($\cdot$): CoT finetune with $\cdot$ steps, WA: With Answer, WoA: Without Answer, OA: Overall Accuracy.}
\label{tab:full-finetune-3tier}
\end{table*}

\begin{figure}[tp!]
\centering
\includegraphics[width=0.47\textwidth]{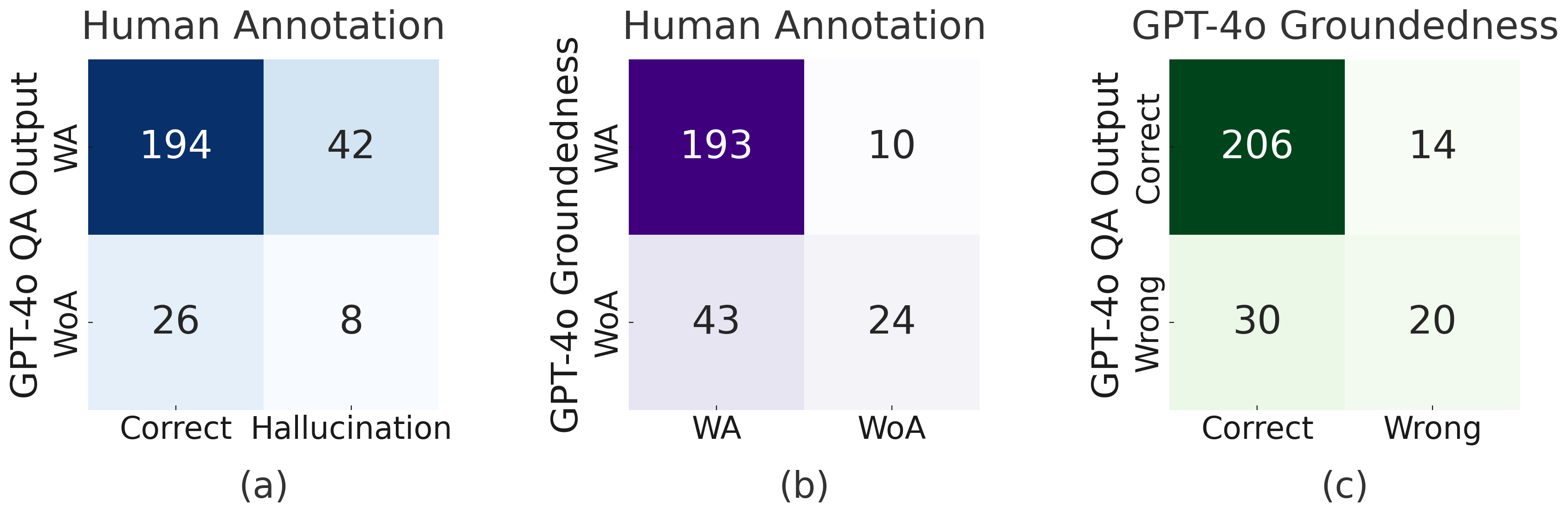}
  \caption{Confusion matrices for evaluation on ESG-Bench. (a) Comparison between GPT-4o's generated answers and human-provided answers.
WA (With Answer) indicates the model produced an answer; WoA (Without Answer) indicates the answer is "Not provided".
(b) Comparison between GPT-4o's binary groundedness judgments (yes/no) and human annotations.
(c) Comparison between GPT-4o's generated answers and its own groundedness judgments (yes/no).
  }
  \label{fig:confusion_matrices}
\end{figure}

\subsubsection{Evaluation for the Generation}

We evaluate all models under a unified WA and WoA paradigm. In the WA setting, the supporting context contains sufficient information to determine the correct answer, the model is expected to read the passage and identify whether the given statement is factually supported. In contrast, in the WoA setting, the correct answer is not provided in the context, meaning that the statement cannot be verified from the passage. This setup enables us to measure whether models will correctly respond with “not provided” rather than hallucinating unsupported claims.
Model outputs are labeled as true or false accordingly. Human annotators verify correctness for ESG-related data, while other datasets rely on gold references under the same WA/WoA definition. We report WA/WoA accuracy, balanced accuracy, and F1 score to capture performance under both conditions. Importantly, GPT-4o is used only for question construction and is not involved in scoring or evaluation, ensuring the objectivity of our results.

\subsection{Experimental Results and Analysis}

\subsubsection{Validating Human Annotations as a Proxy}

To validate the reliability of human-annotated hallucination labels in ESG-Bench, we conduct a proxy evaluation using GPT-4o’s own self-assessment capabilities. This analysis compares: (1) \textit{GPT-4o’s original QA outputs}, (2) \textit{the corresponding human annotations}, and (3) \textit{GPT-4o’s post-hoc yes/no judgment}. As described in the ESG-Bench construction section, we first prompt GPT-4o to answer ESG-related questions using grounded reports. Human annotators then label each answer as correct (WA) or hallucinated (WoA) and provide a reference answer. In a second stage, GPT-4o is re-prompted to assess its own previous outputs using a binary yes/no format: “Given the background, question, and answer, can the answer be found in the background knowledge?”.

Figure~\ref{fig:confusion_matrices} presents three confusion matrices summarizing agreement among these sources. \textit{Subfigure (a)} shows a strong match between GPT-4o’s initial outputs and human annotations (81.5\% agreement). Most answers labeled WA by the model are validated by annotators, while most abstentions align with human-labeled unanswerable cases. Some false abstentions suggest missed opportunities for improved recall. \textit{Subfigure (b)} compares human annotations with GPT-4o’s post-hoc judgments (80.4\% agreement), indicating that GPT-4o can reliably evaluate whether its answers are grounded. \textit{Subfigure (c)} shows strong internal consistency between GPT-4o’s original QA decisions and its later self-evaluations (83.7\%), with most abstentions receiving a “no” and most answers receiving a “yes.”
These results confirm the viability of using GPT-4o’s binary groundedness signal as a proxy supervision for finetuning hallucination-aware models.

\subsubsection{Hallucination Mitigation}

Table~\ref{tab:full-finetune-3tier} presents a comprehensive evaluation of model performance across three datasets (ESG-Bench, HaluEval, and BioASQ), three LLM families (LLaMA, Gemma, and Mistral), and multiple finetuning strategies. Accuracy is reported separately for WA (With Answer) and WoA (Without Answer) cases to fairly assess both answer faithfulness and abstention. This breakdown is critical for hallucination mitigation, as it reflects a model’s dual ability to generate grounded answers and avoid unsupported claims.

This pattern suggests that multi-step reasoning yields a more balanced model, one capable of handling both answerable and unanswerable queries with stability. Such balance is especially important for deployment in domains like ESG, where factual reliability is paramount. These trends also validate our use of GPT-4o’s groundedness judgments (yes/no) as a proxy supervision signal, enabling models to learn from feedback on whether an answer is supported by the context.

Table~\ref{tab:full-finetune-3tier} summarizes these trends using balanced accuracy and F1 scores. These metrics aggregate WA and WoA outcomes to reflect holistic performance, highlighting not only correctness but also the precision-recall tradeoff in abstention behavior. Notably, the 4-step CoT model emerges as the most reliable across datasets and models. High F1 scores in WoA cases demonstrate its capacity to minimize false positives while preserving recall, a key indicator of hallucination mitigation. Together, these findings underscore that CoT finetuning with groundedness-based supervision offers a principled and robust solution to hallucination-aware QA.

Taken together, the results show that CoT finetuning with groundedness-based supervision fundamentally improves how LLMs handle document-grounded QA. Rather than relying on parametric knowledge or shallow pattern matching, models learn to retrieve, filter, 
and verify evidence before producing an output. This makes the 4-step CoT approach a principled and robust solution to hallucination-aware QA in both ESG and general long-context settings.

\section{Conclusion}

As ESG reporting becomes central to corporate accountability and regulation, applying LLMs to these high-stakes documents makes hallucination mitigation essential for ensuring factual reliability. We introduce ESG-Bench, a domain-specific benchmark for evaluating hallucination behavior in long-context ESG question answering, with human-annotated correctness labels and abstention cases to assess both answer faithfulness and conservative reasoning.
Across multiple datasets, models, and training strategies, we show that CoT fine-tuning, particularly with multi-step supervision, substantially improves performance on both answerable and unanswerable queries. We further demonstrate that GPT-4o’s groundedness judgments can serve as an effective proxy supervision signal for hallucination-aware training. Overall, our results indicate that structured reasoning and proxy supervision provide scalable and effective pathways for improving factual reliability in socially sensitive domains.

\section{Ethics Statement}

This study received ethical approval from the University of Sheffield Research Ethics Committee (reference number: 064356). All procedures performed in this study complied with the institutional research ethics standards.

\section{Acknowledgments}

The project was supported by UK’s innovation agency (Innovate UK) grant 10098112 (project name
ASIMOV: AI-as-a-Service).

% \newpage
\bibliography{aaai2026}

\end{document}